%% file: main.tex
\documentclass{article}

\usepackage{booktabs}
\usepackage{microtype}
\usepackage{graphicx}
\usepackage{subfigure}
\usepackage{booktabs} 
\usepackage{hyperref}
\usepackage[accepted]{icml2020}
\usepackage{tablefootnote}
\usepackage{bbold}
\usepackage{mathtools} 
\usepackage{amsmath,amssymb}
\usepackage{enumitem}
\usepackage{xcolor}
\usepackage{caption}
\usepackage{csvsimple}
\usepackage{placeins}
\usepackage{footnote}

\def\reals{\mathbb{R}}

\icmltitlerunning{Revisiting Spatial Invariance with Low-Rank Local Connectivity}

\begin{document}
\definecolor{darkgreen}{rgb}{0,0.6,0}
\newcommand{\todo}[2][]{{\textcolor{darkgreen}{#2}}}

\twocolumn[
\icmltitle{Revisiting Spatial Invariance with Low-Rank Local Connectivity}

\begin{icmlauthorlist}
\icmlauthor{Gamaleldin F. Elsayed}{goo}
\icmlauthor{Prajit Ramachandran}{goo}
\icmlauthor{Jonathon Shlens}{goo}
\icmlauthor{Simon Kornblith}{goo}
\end{icmlauthorlist}

\icmlaffiliation{goo}{Google Research, Brain Team}

\icmlcorrespondingauthor{Gamaleldin F. Elsayed}{gamaleldin@google.com}

\icmlkeywords{Machine Learning, ICML}

\vskip 0.3in
]

\printAffiliationsAndNotice{Author contributions: G.F.E. proposed the project idea. G.F.E. designed the methods with contributions from S.K., P.R., and J.S. G.F.E. wrote the code, conducted experiments, and collected results. S.K. reviewed the code. P.R. conducted FLOP count analysis and designed Figure 2. G.F.E., S.K., P.R., and J.S. wrote the paper.}

\begin{abstract}
Convolutional neural networks are among the most successful architectures in deep learning with this success at least partially attributable to the efficacy of spatial invariance as an inductive bias. Locally connected layers, which differ from convolutional layers only in their lack of spatial invariance, usually perform poorly in practice. However, these observations still leave open the possibility that some degree of relaxation of spatial invariance may yield a better inductive bias than either convolution or local connectivity. To test this hypothesis, we design a method to relax the spatial invariance of a network layer in a controlled manner; we create a \textit{low-rank} locally connected layer, where the filter bank applied at each position is constructed as a linear combination of basis set of filter banks with spatially varying combining weights. By varying the number of basis filter banks, we can control the degree of relaxation of spatial invariance. In experiments with small convolutional networks, we find that relaxing spatial invariance improves classification accuracy over both convolution and locally connected layers across MNIST, CIFAR-10, and CelebA datasets, thus suggesting that spatial invariance may be an overly restrictive prior.

\end{abstract}

\input{introduction}
\input{related_work}
\input{methods}

\input{results}

\input{conclusion}

\section{Acknowledgements}
We are grateful to Jiquan Ngiam, Pieter-Jan Kindermans, Jascha Sohl-Dickstein, Jaehoon Lee, Daniel Park, Sobhan Naderi, Max Vladymyrov, Hieu Pham, Michael Simbirsky, Roman Novak, Hanie Sedghi, Karthik Murthy, Michael Mozer, and Yani Ioannou for useful discussions and helpful feedback on the manuscript. 
\FloatBarrier
\bibliography{main}
\bibliographystyle{icml2020}
\input{appendix}

\end{document}

%% file: introduction.tex
\section{Introduction}
\label{introduction}

\begin{figure}[t!]
\vskip 0.2in
\begin{center}
\centerline{\includegraphics[width= \columnwidth]{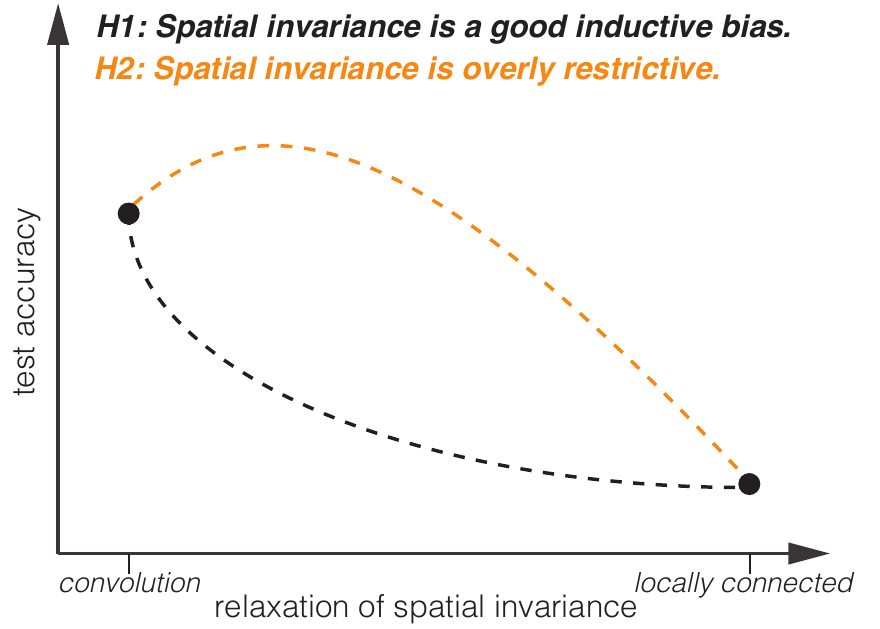}}
\caption{\textbf{Is spatial invariance a good inductive bias?} Convolutional architectures perform better than locally connected (or fully connected) architectures on computer vision problems. The primary distinction between convolutional and locally connected networks is requiring spatial invariance in the learned parameter set. Spatial invariance is imposed through weight sharing. One long-standing hypothesis (H1) is that this spatial invariance is a good inductive bias for images \cite{ruderman1994statistics,simoncelli2001natural,olshausen1996natural}. H1 posits that predictive performance would systematically degrade as spatial invariance is relaxed. An alternative hypothesis (H2) suggests that spatial invariance is overly restrictive and some degree of variability would aid predictive performance. The degree to which H1 or H2 is a good hypothesis is largely untested across natural and curated academic datasets and the subject of this work.}
\label{fig:cartoon}
\end{center}
\vskip -0.2in
\end{figure}

\begin{figure*}[t!]
\vskip 0.2in
\begin{center}
\centerline{\includegraphics[width= \columnwidth*2]{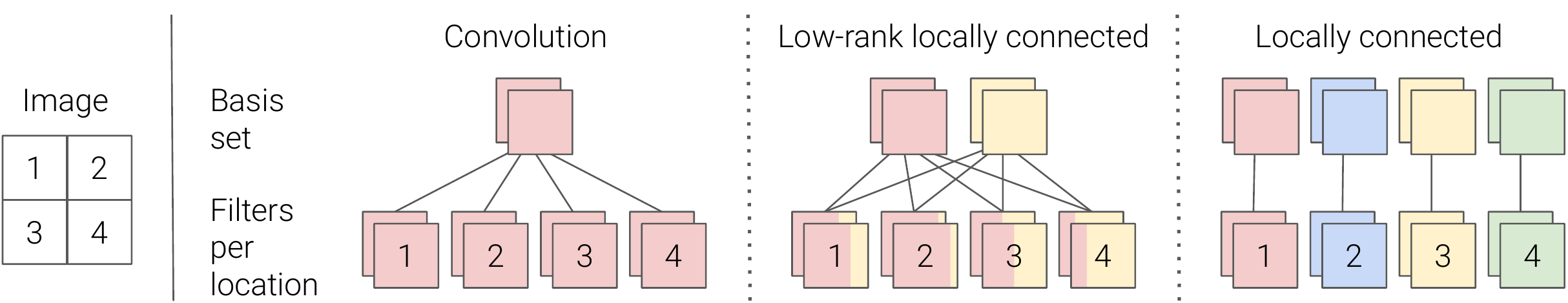}}
\caption{\textbf{Filters for each spatial location.} Convolutional layers use the same filter bank for each spatial location (left). Locally connected layers learn a separate filter bank for each spatial location (right). By contrast, low-rank locally connected (LRLC) layers use a filter bank for each spatial location generated from combining a shared basis set of filter banks (middle). Both the basis set and the combining weights are learned end-to-end through optimization. The number of filter banks in the basis set (i.e., the rank parameter) thus determines the degree of relaxation of spatial invariance of the LRLC layer.}
\label{fig:kernel_generation}
\end{center}
\vskip -0.2in
\end{figure*}

Convolutional neural networks (CNNs) are now the dominant approach across many computer vision tasks. 
Convolution layers possess two main properties that are believed to be key to their success: local receptive fields and 
spatially invariant filters.
In this work, we seek to revisit the latter.
Previous work comparing convolutional layers, which share filters across all spatial locations, with locally connected layers, which have no weight sharing, has found that 
convolution is advantageous on common datasets~\cite{lecun-89,bartunov2018assessing,novak2018bayesian}. However, this observation leaves open the possibility that some departure from spatial invariance could outperform both convolution and local connectivity (Figure~\ref{fig:cartoon}).

The structure of CNNs is often likened to the primate visual system~\cite{lecun2015deep}. However, the visual system has no direct mechanism to share weights across space. Neurons comprising retinotopic maps have selectivity properties that vary with their position within the map, particularly in high-level visual areas~\cite{hasson2002eccentricity,arcaro2009retinotopic,lafer2013parallel,rajimehr2014retinotopy,srihasam2014novel,saygin2016connectivity,livingstone2017development}. Moreover, the retina contains several types of cells whose distribution and features are organized according to low-rank spatial gradients~\cite{dacey1992dendritic}.

Motivated by the lack of synaptic weight sharing in the brain,
we hypothesized that neural networks could achieve greater performance by relaxing spatial invariance (Figure \ref{fig:cartoon}). 
Particularly at higher layers of the neural network, where receptive fields cover most or all of the image, applying the same weights at all locations may be a less efficient use of computation than applying different weights at different locations. However, evidence suggests that typical datasets are too small to constrain the parameters of a locally connected layer; functions expressible by convolutional layers are a subset of those expressible by locally connected layers, yet convolution typically achieves higher performance~\cite{lecun-89,bartunov2018assessing,novak2018bayesian}.  

To get intuition for why some relaxation of spatial invariance could be useful, consider images of natural scenes with ground and sky regions. It may be a bad idea to apply different local filters to different parts of the sky with similar appearance. However, it may also be overly limiting to apply the same filter bank to both the sky and the ground regions. Some degree of relaxation of the spatial invariance, such as a different sky and ground filters, may better suit this hypothetical data.

To test the hypothesis that spatial invariance is an overly restrictive inductive bias, we create a new tool that allows us to relax spatial invariance. 
We develop a \textit{low-rank} locally connected (LRLC) layer\footnote{Code is available at \href{https://github.com/google-research/google-research/tree/master/low_rank_local_connectivity}{github.com/google-research/google-research/tree/master/low\textunderscore rank\textunderscore local\textunderscore connectivity}.} that can parametrically adjust the degree of spatial invariance. 
This layer is one particular method to relax spatial invariance by reducing weight sharing.
Rather than learning a single filter bank to apply at all positions, as in a convolutional layer, or different filter banks, as in a locally connected layer, the LRLC layer learns a set of $K$ filter banks, which are linearly combined using $K$ combining weights per spatial position (Figure~\ref{fig:kernel_generation}).

In our experiments,
we find that relaxing spatial invariance with 
the LRLC layer leads to better performance compared to both convolutional and locally connected layers across three datasets (MNIST, CIFAR-10, and CelebA). These results suggest that some level of relaxation of spatial invariance is a better inductive bias for image datasets compared to the spatial invariance enforced by convolution layers or lack of spatial invariance in locally connected layers.

%% file: related_work.tex
\section{Related Work}
\label{related_work}

The idea of local connectivity in connectionist models predates the popularity of backpropagation and convolution.
Inspired by the organization of visual cortex \cite{hubel1963shape,hubel1968receptive}, several early neural network models consisted of one or more two-dimensional feature maps where neurons preferentially receive input from other neurons at nearby locations \cite{von1973self,fukushima1975cognitron}.
Breaking with biology, the Neocognitron~\cite{fukushima1980neocognitron} shared weights across spatial locations, resulting in spatial invariance. However, the Neocognitron was trained using a competitive learning algorithm rather than gradient descent. \citet{lecun-89} combined weight sharing with backpropagation, demonstrating considerable gains over locally connected networks (LCNs) on a digit recognition task.

Although the last decade has seen revitalized interest in CNNs for computer vision, local connectivity has fallen out of favor.
When layer computation is distributed across multiple nodes, weight sharing introduces additional synchronization costs \cite{krizhevsky2014one}; thus, the first massively parallel deep neural networks employed exclusively locally connected layers \cite{raina2009large,uetz2009large,dean2012large,le2012building,coates2013deep}. Some of the first successful neural networks for computer vision tasks combined convolutional and locally connected layers~\cite{hinton2012improving,goodfellow2013multi,gregor2013deep}, as have networks for face recognition~\cite{taigman2014deepface,sun2014deep,sun2015deepid3,yim2015rotating}.  However, newer architectures, even those designed for face recognition~\cite{schroff2015facenet,liu2017sphereface}, generally use convolution exclusively.

Work comparing convolutional and locally connected networks for computer vision tasks has invariably found that CNNs yield better performance. \citet{bartunov2018assessing} compared the classification performance on
multiple image datasets as part of a study on biologically plausible learning algorithms; convolution achieved higher accuracy across datasets. \citet{novak2018bayesian} derived a kernel equivalent to an infinitely wide CNN at initialization and showed that, in this infinite-width limit, CNNs and LCNs are equivalent. They found that SGD-trained CNNs substantially outperform both SGD-trained LCNs and this kernel. However, \citet{d2019finding} found that initially training a convolution layer and then converting the convolutional layers to equivalent fully connected layers near the end of training led to a slight increase in performance.

Other work has attempted to combine the efficiency of convolution with some of the advantages of local connectivity. \citet{nowlan1992simplifying} suggested a ``soft weight-sharing" approach that penalizes the difference between the distribution of weights and a mixture of Gaussians. Other work has used periodic weight sharing, also known as tiling, where filters $n$ pixels away share weights~\cite{le2010tiled,gregor2010emergence}, or subdivided feature maps into patches where weights are shared only within each patch~\cite{zhao2016deep}. CoordConv \cite{liu2018intriguing} concatenates feature maps containing the $x$ and $y$ coordinates of the pixels to the input of a CNN, permitting direct use of position information throughout the network.

Input-dependent low rank local connectivity, which we explore in Sections~ \ref{sec:input-dependent LRLC} and \ref{sec:input_dependent}, is further related to previous work that applies input-dependent convolutional filters. Spatial soft attention mechanisms~\cite{wang2017residual,jetley2018learn,woo2018cbam,linsley2018learning,fukui2019attention} can be interpreted as a mechanism for applying different weights at different positions via per-position scaling of entire filters. Self-attention~\cite{bahdanau2015neural,vaswani2017attention}, which has recently been applied to image models~\cite{bello2019attention,ramachandran2019stand,hu2019local}, provides an alternative mechanism to integrate information over space with content-dependent mixing weights. Non-local methods \citep{wang2018non,zhang2019latentgnn} and graph convolution approaches \citep{chen2019graph} are additional ways to perform content-dependent spatial aggregation.
Other approaches apply the same convolutional filters across space, but select filters or branches separately for each example~\cite{mcgill2017deciding,fernando2017pathnet,gross2017hard,chen2019you,yang2019condconv}. 
The dynamic local filtering layer of~\citet{jia2016dynamic} uses a neural network to predict a separate set of filters for each position. Our approach predicts only the combining weights for a fixed set of bases, which provides 
control over the degree of spatial invariance through the size of the layer kernel basis set. The CondConv layer of \citet{yang2019condconv} predicts combining weights per example that are shared across all spatial locations, whereas our approach learns weights per spatial location, optionally dependent on the example. Further, the computation of spatial filters in the input-dependent LRLC layer can be thought of as a form of dynamic routing, which relates to Capsule networks \citep{sabour2017dynamic}. However, in \citet{sabour2017dynamic}, the first capsule layer (PrimaryCaps) is convolutional and fully connects to every DigitCaps capsule, which does not allow partial relaxation of spatial invariance as in the LRLC layer.

%% file: methods.tex
\section{Methods}
\label{methods}

\subsection{Preliminaries}

Let $I \in  \reals^{H \times W \times C_{\text{in}}}$ be an input with $C_{\text{in}}$ channels ($H$: input height, $W$: input width, and $C_{in}$: input channels). In convolution layers, the input $I$ is convolved with a filter bank $F \in \reals^{h \times w \times C_{\text{in}} \times C_{\text{out}}}$ to compute $O \in \reals^{H \times W \times C_{\text{out}}}$ ($h$: filter height size, $w$: filter width size, and $C_{out}$: filter output channels).
For clarity of presentation, we fix the layer output and input to have the same size, and the stride to be 1, though we relax these constraints in the experiments. More formally, the operation of $F$ on the local input patch of size $h \times w \times C_{\text{in}}$ centered at location $(i, j)$, $I_{i,j}$, is:
\begin{align}
O_{i,j} = I_{i,j} \star F \stackrel{\mathrm{def}}{=} \sum_{x=1}^{h} \sum_{y=1}^{w} \sum_{z=1}^{C_{\text{in}}} \left( I_{i,j} \odot F \right)_{x,y,z} 
\end{align}
where $O_{i,j} \in \reals^{C_{\text{out}}}$ is the output
at location $(i, j)$ $\forall i \in \{1,\dots, H \}$ and $\forall j \in \{1,\dots, W \}$ ($\odot$ is defined as the element-wise multiplication of the input and the filter along the first 3 axes). The spatial invariance of convolution refers to applying the same filter bank $F$ to input patches at all locations (Figure \ref{fig:kernel_generation} left). 

Locally connected layers on the other hand do not share weights. Similar to convolution, they apply filters with local receptive fields. However, the filters are not shared across space (Figure \ref{fig:kernel_generation} right). Formally, each output $O_{i,j}$ is computed by applying a different filter bank $F^{(i,j)}$ to the corresponding input patch (i.e., $O_{i,j} = I_{i,j} \star F^{(i,j)}$). 

Empirically, locally connected layers perform poorly compared to convolutional layers \citep{novak2018bayesian}. Intuitively, local regions in images are not completely independent and we expect filters learned over one local region to be useful when applied to a nearby region. While locally connected layers are strictly more powerful than convolutional layers and could {\em in theory} converge to the convolution solution, in practice 
they don't and instead overfit the training data. However, the superior performance of convolution layers over locally connected layers \cite{lecun-89,bartunov2018assessing,novak2018bayesian} does not imply that 
spatial invariance is strictly required.

Below, we develop methods that control the degree of spatial invariance a layer can have, which allows us to test the hypothesis that spatial invariance may be overly restrictive.

\subsection{Low-rank locally connected layer}

Here, we design a locally connected layer with a spatial rank parameter that controls the degree of spatial invariance. We adjust the degree of spatial invariance by using a set of $K$ local filter banks (\emph{basis set}) instead of $1$ filter bank in a convolution layer or $H \times W$ filter banks in a classic locally connected layer ($K$ is a hyperparameter that may be adjusted based on a validation subset; $1 \le K \le H\times W$). For each input patch $I_{i,j}$, we construct a filter bank to operate on that patch that is a linear combination of the members of the basis set. That is, 
\begin{align}
F^{(i,j)} = \sum_{k=1}^{K} w_{i,j}^{(k)} F^{(k)}
\end{align}
where $w_{i,j}^{(k)} \in \reals$ are the weights that combine the filter banks in the basis set $\forall i \in \{1,\dots, H \}$ and $\forall j \in \{1,\dots, W \}$. This formulation is equivalent to a low-rank factorization with rank $K$ of the layer locally connected kernel. Thus, we term this layer the ``low-rank locally connected" (\emph{LRLC}) layer (Figure \ref{fig:kernel_generation} middle).

Note that, in 
this paper, we use basis set with filters of similar structure. However, this layer could also be used with a basis set containing filters with different structure (e.g., different filter sizes and/or dilation rates). 

The filters in the basis set are linearly combined using weights that are specific to each spatial location. In particular, with input size $H\times W$ and $K$ filter banks in the basis set, we need $H \times W \times K$ weights to combine these filter banks and formulate the filter bank at each spatial location. We propose two ways to learn these combining weights. One method learns weights that are shared across all examples while the second method 
predicts the weights per example based on a function of the input.

\subsubsection{Fixed combining weights}
\label{sec: Fixed kernel combining weights}

The simplest method of learning combining weights is to learn $K$ scalars per spatial position. This approach is well-suited to datasets with spatially inhomogeneous features, \textit{e.g.} datasets of aligned faces. The number of combining weights scales linearly with the number of pixels in the image, which may be large. Thus to reduce parameters, we learn combining weights per-row and per-column 
of location $(i, j)$ as follows:
\begin{align}
\label{eqn: combining weights}
\tilde{w}_{i, j}^{(k)} = \alpha_{i}^{(k)} + \beta_{j}^{(k)}
\end{align}
This formulation reduces the number of combining weights parameters to $(H+W) \times K$, which limits the expressivity of the layer (i.e., constrains the maximum degree of relaxation of spatial invariance). This formulation also performs better in practice (Figure \ref{fig: factorized weights and biases}).

We further normalize the weights to limit the scale of the combined filters. Common choices for normalization are dividing by the weights norm or using the softmax function. In our early experimentation, we found that softmax normalization performs slightly better. Thus, the combining weights are computed as follows:
\begin{align}
\label{eqn: combining weights normalized}
w_{i, j}^{(k)} = \frac{\text{exp}\left(\tilde{w}_{i,j}^{(k)}\right)}{\sum_{l=1}^K \text{exp}\left(\tilde{w}_{i,j}^{(l)}\right)} 
\end{align}

The filter banks in the basis set and the combining weights can all be learned end-to-end. In practice, we implement this layer with convolution and point-wise multiplication operations, as in Algorithm \ref{alg: layer}, rather than forming the equivalent locally connected layer. This implementation choice is due to locally connected layers being slower in practice because current hardware is memory-bandwidth-limited, while convolution is highly optimized and fast.
We initialize the combining weights to a constant, which is equivalent to a  convolution layer with a random kernel, though our main findings remained the same with or without this initialization (Figure \ref{fig: structured initialization}).

At training time, the parameter count of the LRLC layer is approximately $K$ times that of a corresponding convolutional layer, as is the computational cost of Algorithm~\ref{alg: layer}. However, after the network is trained, the LRLC layer can be converted to a locally connected layer. When convolution is implemented as matrix multiplication, locally connected layers have the same FLOP count as convolution (Figure \ref{fig: lrlc flops}), although the amount of memory needed to store the weights scales with the spatial size of the feature map.

\begin{algorithm}[tb]
\caption{Low Rank Locally Connected Layer}
\label{alg: layer}
\begin{algorithmic}
    \STATE {\bfseries Input:} $I \in \reals^{H \times W \times C_{\text{in}}}$
    \STATE {\bfseries Trainable Parameters:} \\
    $\{F^{(1)}, \dots, F^{(K)}\} \in K \reals^{h \times w \times C_{\text{in}}  \times C_{\text{out}}}$ (basis set)
    
    $\alpha^{(1)}, \dots, \alpha^{(K)} \in \reals^{H}$ (combining weights for rows)
    
    $\beta^{(1)}, \dots, \beta^{(K)} \in \reals^{W}$ (combining weights for columns)
    
    $b^{\text{row}} \in \reals^{H}$ (biases for rows)
    
    $b^{\text{column}} \in \reals^{W}$ (biases for columns)
    
    $b^{\text{channel}} \in \reals^{C_{\text{out}}}$ (biases for channels)
    
    Initialize $O = 0 \in \reals^{H \times W \times C_{out}}$.

    \FOR{$k=1$ {\bfseries to} $K$}
       \item $O^{(k)} = I \circledast F^{(k)}$  (convolution with filters in basis set)
       \item  $W^{(k)} = \alpha^{(k)} 1_{W}^\top + 1_{H} {\beta^{(k)}}^\top$ (combining weights)
       \item $O = O + W^{(k)} \odot O^{(k)}$
    \ENDFOR
    
   $B_{i,j,c} = b^{\text{row}}_i + b^{\text{column}}_j + b^{\text{channel}}_c$ (biases)
   
   $O = O + B$ 
   \STATE{\bfseries Return:} O
\end{algorithmic}
\end{algorithm}

\paragraph{Spatially varying bias} Typically, a learned bias per channel is added to the output of a convolution. Here, we allow the bias that is added to the LRLC output to also vary spatially. Similar to the combining weights, per-row and per-column biases are learned and are added to the standard channel bias. Formally, we define the layer biases ($B$) as:
\begin{align}
\label{eqn: biases}
B_{i,j,c} = b^{\text{row}}_i + b^{\text{column}}_j + b^{\text{channel}}_c
\end{align}
where $b^{\text{row}} \in \reals^{H}$,  $b^{\text{column}} \in \reals^{W}$,  and $b^{\text{channel}} \in \reals^{C_{\text{out}}}$. The special case of the LRLC layer with $K = 1$ is equivalent to a convolution operation followed by adding the spatially varying bias. We use this case in our experiments as a simple baseline to test if relaxing spatial invariance in just the biases is enough to see improvements.

\begin{table*}[!h]
 \caption{\textbf{Spatial invariance may be overly restrictive.} Top-1 accuracy of different models (mean $\pm$ SE).
 The optimal rank in LRLC is obtained by evaluating models on a separate validation subset.}
 \label{table1}
\vskip 0.1in
\begin{center}
\begin{small}
\begin{sc}
\begin{tabular}{lcccr}
\begin{filecontents*}{figures/results_table1.csv}
Layer,MNIST,CIFAR-10,CelebA
Convolution,$98.84\pm0.01$,$78.80\pm0.12$,$96.66\pm0.04$
Convolution + spatial bias (1st layer),$99.37\pm0.01$,$80.94\pm0.10$,$97.08\pm0.02$
Convolution + spatial bias (2nd layer),$99.25\pm0.02$,$80.60\pm0.05$,$97.06\pm0.03$
Convolution + spatial bias (3rd layer),$99.07\pm0.02$,$79.94\pm0.12$,$96.76\pm0.02$
Convolution + spatial bias (all layers),$99.36\pm0.01$,$80.89\pm0.08$,$97.12\pm0.03$
LRLC (1st layer),$99.39\pm0.02$,$80.94\pm0.10$,$97.25\pm0.02$
LRLC (2nd layer),$99.42\pm0.02$,$80.67\pm0.09$,$97.53\pm0.01$
LRLC (3rd layer),$\boldsymbol{99.51\pm0.01}$,$\boldsymbol{82.70\pm0.08}$,$\boldsymbol{97.73\pm0.02}$
LRLC (all layers),$99.45\pm0.01$,$81.03\pm0.11$,$97.51\pm0.03$
\end{filecontents*}
\csvautobooktabular{figures/results_table1.csv}
\end{tabular}
\end{sc}
\end{small}
\end{center}
\vskip -0.1in
\end{table*}

\begin{figure*}[t!]
\vskip 0.2in
\begin{center}
\centerline{\includegraphics[width= \columnwidth*2]{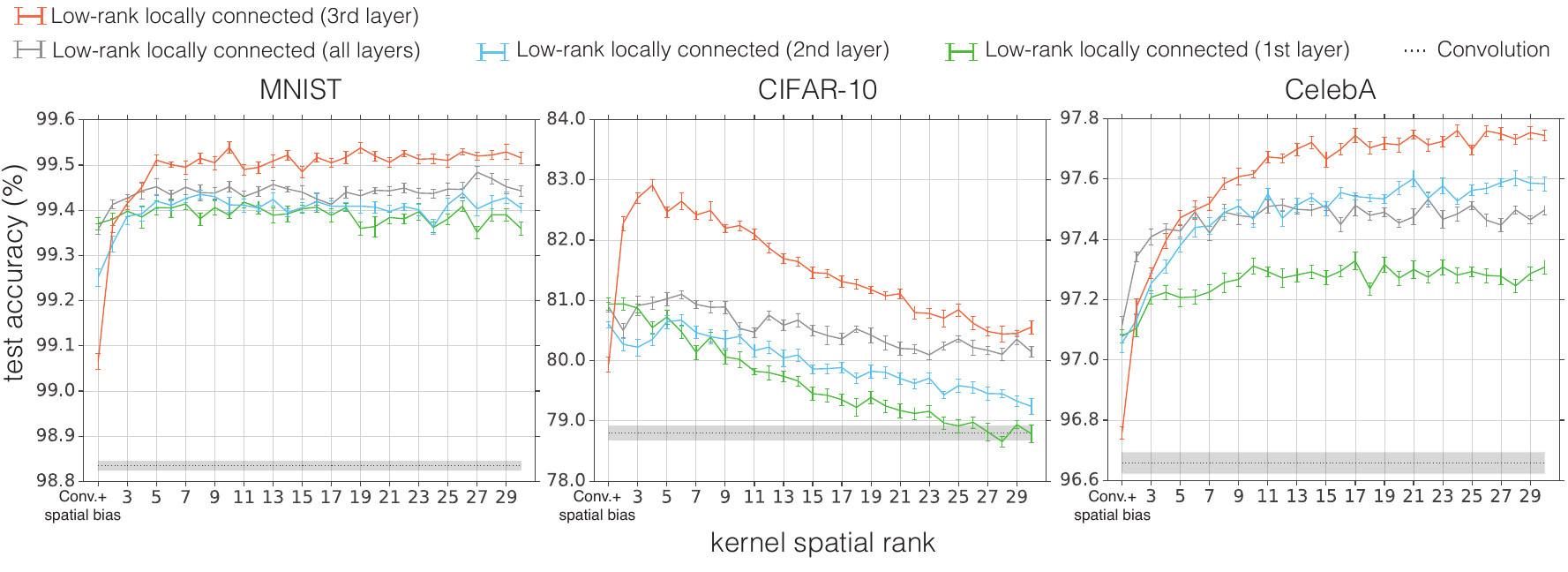}}
\caption{\textbf{Low-rank local connectivity is a good inductive bias for image datasets.} 
Vertical axis shows top-1 test accuracy on digit classification task on $28 \times 28$ images from MNIST dataset (left), object classification task on $32 \times 32$ images from CIFAR-10 dataset (middle), and gender classification task on $32 \times 32$ images from CelebA dataset (right). Horizontal axis shows the locally connected kernel spatial rank used for the low-rank locally connected (LRLC) layer placed at first, second, third, layer or all layers of the network. Note, we only includes low ranks ($\leq 30$) due to computational constraints. An LRLC layer would require rank $32^2=1024$ for a CIFAR-10 for example to match the effective rank of a locally connected layer. The accuracy of a regular convolutional network is shown as a dotted black line for reference. Error bars indicate $\pm$ standard errors computed from training models from 10 random initialization. The LRLC layer outperforms classic convolution, suggesting that convolution is overly restrictive and consistent with H2 in Figure \ref{fig:cartoon}.}
\label{fig: lrlc}
\end{center}
\vskip -0.2in
\end{figure*}

\subsubsection{Input-Dependent Combining Weights}
\label{sec:input-dependent LRLC}

The fixed combining weights formulation intuitively will work best when all images are aligned with structure that appears consistently in the same spatial position. Many image datasets have some alignment by construction, and we expect this approach to be particularly successful for such datasets. However, this formulation may not be well-suited to datasets without image alignment.
In this section, we describe an extension of the LRLC layer that conditions the combining weights on the input.

Formally, we modify the combining weights in equation \ref{eqn: combining weights} to make them a function of the input:
\begin{align}
\label{eqn: dynamic combining weights}
\tilde{w}_{i, j}^{(k)} = g_{i,j}^{(k)}(I)
\end{align}
where $g$ is a lightweight neural network that predicts the combining weights for each position. More formally, $g$ takes in the input $I \in \reals^{H \times W \times C_{\text{in}}}$ and outputs weights $\tilde{w} \in \reals^{H \times W \times K}$. The predicted weights are then similarly normalized as in equation \ref{eqn: combining weights normalized} and are used as before to combine the filter banks in the basis set to form local filters for each spatial location. Similar to section \ref{sec: Fixed kernel combining weights}, a spatially varying bias is also applied to the output of the layer. The architecture used for $g$ has low computational cost, consisting of several dilated separable convolutions applied in parallel followed by a small series of cheap aggregation layers that output a $H \times W \times K$ tensor. The full architecture of $g$ is detailed in the supplementary section \ref{sec:input-dependent-kernel-network} and shown in Figure \ref{fig: input-dependent network}.

%% file: results.tex
\section{Experiments}
\label{experiments}

\begin{table*}[t]
 \caption{\textbf{LRLC outperforms baselines.} Top-1 accuracy of different models (mean $\pm$ SE). The optimal rank for LRLC and the optimal width for wide convolution models are obtained by evaluating models on a separate validation subset.}
 \label{table2}
\vskip 0.1in
\begin{center}
\begin{small}
\begin{sc}
\begin{tabular}{lcccr}
\begin{filecontents*}{figures/results_table2.csv}
Layer,MNIST,CIFAR-10,CelebA
LRLC (3rd layer),$\boldsymbol{99.51\pm0.01}$,$\boldsymbol{82.70\pm0.08}$,$\boldsymbol{97.73\pm0.02}$
CoordConv \cite{liu2018intriguing},$99.46\pm0.01$,$81.29\pm0.13$,$97.40\pm0.02$
Locally connected (1st layer),$98.74\pm0.01$,$62.54\pm0.14$,$96.32\pm0.02$
Locally connected (2nd layer),$98.72\pm0.02$,$69.29\pm0.09$,$97.19\pm0.02$
Locally connected (3rd layer),$99.10\pm0.01$,$71.86\pm0.10$,$97.31\pm0.01$
Wide convolution (3rd layer),$99.10\pm0.02$,$82.40\pm0.10$,$97.20\pm0.02$
\end{filecontents*}
\csvautobooktabular{figures/results_table2.csv}
\end{tabular}
\end{sc}
\end{small}
\end{center}
\vskip -0.1in
\end{table*}

\begin{figure*}[t!]
\vskip 0.2in
\begin{center}
\centerline{\includegraphics[width= \columnwidth*2]{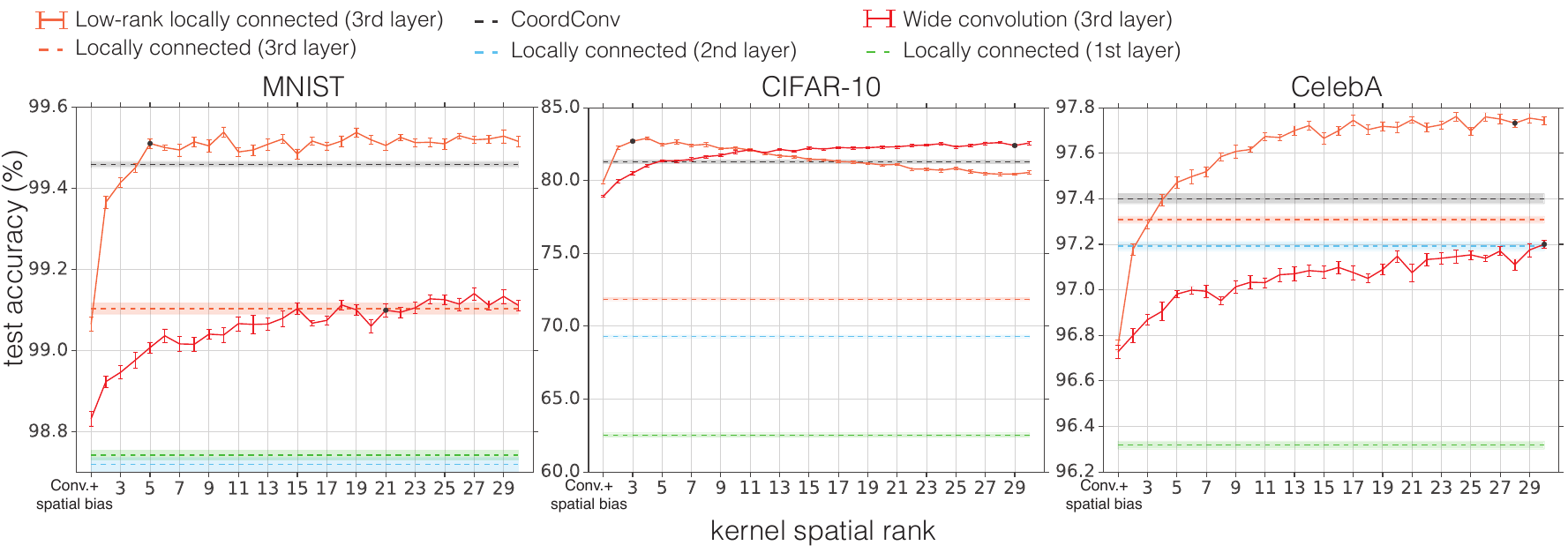}}
\caption{\textbf{LRLC outperforms baselines.} Similar to Figure \ref{fig: lrlc}, comparing the LRLC layer to different baselines.
Baselines include standard locally connected layers, CoordConv \cite{liu2018intriguing}, and convolution networks with wider channels than 64 with width adjusted to match the number of parameters in the LRLC layer. Black markers indicate the best model across spatial ranks for LRLC models and across different widths for wide convolution models. The best models are obtained by performing evaluation on a separate validation subset.}
\label{fig: lrlc_baselines}
\end{center}
\vskip -0.2in
\end{figure*}
We performed classification experiments on MNIST, CIFAR-10, and CelebA datasets. We trained our models without data augmentation or regularization to focus our investigation on the pure effects of the degree of spatial invariance on generalization. In our experiments, we used the Adam optimizer with a maximum learning rate of $0.01$ and a minibatch size  of $512$. We trained our models for $150$ epochs starting with a linear warmup period of $10$ epochs and used a cosine decay schedule afterwards. We used Tensor Processing Unit (TPU) accelerators in all our training.

We conducted our study using a network of $3$ layers with $64$ channels at each layer and local filters of size $3\times 3$. Each layer is followed by batch normalization and ReLU nonlinearity. The network is followed by a global average pooling operation then a linear fully connected layer to form predictions. Our network had sufficient capacity, and we trained for sufficiently large number of steps to 
achieve high training accuracy (Table \ref{table: train accuracy}). For all our results, we show the mean accuracy $\pm$ standard error based on models trained from 10 different random initializations. Our division of training, validation and test subsets are shown in Table \ref{table: data subsets}.

\subsection{Spatial invariance may be overly restrictive}

In this section, we investigate whether relaxing the degree of spatial invariance of a layer is a better inductive bias for image classification. We replaced convolution layers at different depths of the network (first, second, third or at all layers) with the designed low-rank locally connected (LRLC) layer. We varied the spatial rank of the LRLC layer, which controls the deviation degree from spatially invariant convolution layers towards locally connected layers. If the rank is small the network is constrained to share filters more across space and the higher the rank the less sharing is imposed. We trained our models and quantified the generalization accuracy on test data at these different ranks.

When rank is 1, the LRLC layer is equivalent to a convolution layer with an additional spatial bias. Adding this spatial bias to the convolution boosted the accuracy over normal convolution layers (Table \ref{table1}). Increasing the spatial rank allows the layer to use different filters at different spatial locations, and deviate further from convolution networks. Our results show that doing so further increases accuracy (Figure \ref{fig: lrlc}). We find that accuracy of networks with LRLC layers placed at any depth, or with all layers replaced by LRLC layers, is higher than that of pure convolutional networks (Figure \ref{fig: lrlc} and Table \ref{table1}). These findings provide evidence for the hypothesis that spatial invariance may be overly restrictive. Our results further show that relaxing the spatial invariance late in the network (near the network output) is better than early (at the input). Relaxing the spatial invariance late in the network was also better than doing so at every layer (Table \ref{table1}).
The optimal spatial rank varied across different datasets; rank was the lowest for CIFAR-10 data and was the highest for CelebA.

The LRLC layer has the ability to encode position, which vanilla convolution layers lack. This additional position encoding may explain the increased accuracy. Previous work has attempted to give this capability to convolution networks by augmenting the input with coordinate channels, an approach known as CoordConv \cite{liu2018intriguing}. 
To test whether the efficacy of the LRLC layer could be explained solely by its ability to encode position, we compared its performance to that of CoordConv.
Our results show that CoordConv outperforms vanilla convolution, but still lags behind the LRLC network (Table \ref{table2} and Figure \ref{fig: lrlc_baselines}), suggesting that the inductive bias of the LRLC layer is better-suited to the data. Unlike CoordConv, the LRLC layer allows controlling and adapting the degree of spatial invariance to different datasets by adjusting the spatial rank. However, with CoordConv, this adjustment is not possible. This gives an intuition of why the LRLC layer suits the data better than CoordConv.

\begin{table*}[t]
 \caption{\textbf{Fixed vs input-dependent combining weights.} Top-1 accuracy of different models (mean $\pm$ SE). The optimal rank is obtained by evaluating models on a separate validation subset.}
 \label{table3}
\vskip 0.1in
\begin{center}
\begin{small}
\begin{sc}
\begin{tabular}{lcccr}
\begin{filecontents*}{figures/results_table3.csv}
Layer,MNIST,CIFAR-10,CelebA
LRLC (3rd layer \& input dependent weights),$\boldsymbol{99.57\pm0.01}$,$\boldsymbol{84.91\pm0.07}$,$97.49\pm0.03$
LRLC (3rd layer),$99.51\pm0.01$,$82.70\pm0.08$,$\boldsymbol{97.73\pm0.02}$
\end{filecontents*}
\csvautobooktabular{figures/results_table3.csv}
\end{tabular}
\end{sc}
\end{small}
\end{center}
\vskip -0.1in
\end{table*}

\begin{figure*}[t!]
\vskip 0.2in
\begin{center}
\centerline{\includegraphics[width=\columnwidth*2]{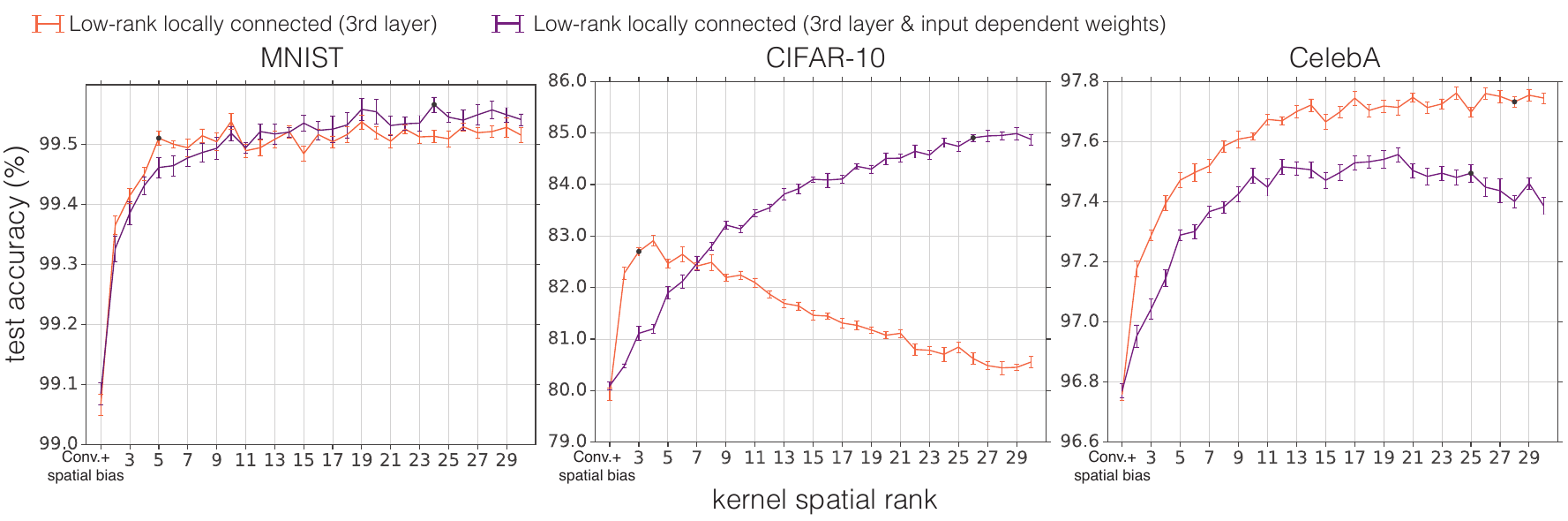}}
\caption{\textbf{Input-dependent combining weights.} 
The LRLC layer learns fixed weights to combine filter banks in the basis set and construct a filter bank to be applied to each input location. The input-dependent LRLC layer uses a simple network to adapt the combining weights to different inputs, making it more suitable for less aligned data such as CIFAR-10. The accuracy of the input-dependent LRLC layer substantially exceeds the accuracy of the fixed LRLC layer on CIFAR-10. However, for more spatially aligned datasets such as MNIST and CelebA, input-dependent LRLC yields modest or no improvement.}
\label{fig: fixed vs input dependent ranks}
\end{center}
\vskip -0.2in
\end{figure*}

Although locally connected layers have inference-time FLOP count similar to standard convolution layers, the relaxation of spatial invariance comes at the cost of an increase number of trainable parameters. In particular, the number of trainable parameters in the LRLC layer grows linearly with the spatial rank (ignoring the combining weights and spatial biases as they are relatively small).
This increase in model parameters does not explain the superiority of the LRLC layer. Locally connected layers have more trainable parameters than LRLC layers, yet perform worse (Figure \ref{fig: lrlc_baselines} and Table \ref{table2}). Moreover, even after widening convolutional layers to match the trainable parameter count of the LRLC layer, networks with only convolutional layers still do not match the accuracy of networks with low-rank locally connected layers (Figures \ref{fig: lrlc_baselines}, \ref{fig: fixed vs input dependent params} and Table \ref{table2}). Thus, in our experiments, LRLC layers appear to provide a better inductive bias independent of parameter count.

\subsection{Input-dependent low-rank local connectivity is a better inductive bias for datasets with less alignment}
\label{sec:input_dependent}

In the previous section, our results show that the optimal spatial rank is dataset-dependent. The spatial rank with highest accuracy (the optimal rank) was different across datasets and was generally far from the full rank (i.e., the spatial size of the input), which gives an intuition why convolution layers work well on images in the context of convolution being closer to the optimal rank compared to the vanilla locally connected layers. The optimal rank seems to depend on alignment in the dataset. For example, the optimal rank was highest for CelebA dataset, which comprises approximately aligned face images. By contrast, on CIFAR-10, the optimal rank was low, which may reflect the absence of alignment in the dataset beyond a weak bias toward objects in the center of the images.

These findings raise the question whether one can achieve more gains if the allocation of local filters across space was not fixed across the whole dataset, but rather was conditioned on the input. To answer this question, we modified the LRLC layer to allow the layer to assign local filters based on the input (see Section \ref{sec:input-dependent LRLC}). This approach has some resemblance to previous work on input-dependent filters \cite{yang2019condconv, jia2016dynamic}. We tested whether using this input-dependent way of selecting local filters can give more gains in the less aligned CIFAR-10 dataset. Our results show that the input-dependent LRLC network indeed achieves higher accuracy on CIFAR-10 compared to the fixed LRLC layer, and yields a higher optimal spatial rank (Figure \ref{fig: fixed vs input dependent ranks} and Table \ref{table3}). We also experimented the input-dependent LRLC on MNIST and CelebA. We found that the input-dependent LRLC only helped a little on MNIST and hurt accuracy on CelebA compared to the LRLC with fixed weights (Figure \ref{fig: fixed vs input dependent ranks} and Table \ref{table3}). This finding suggests that the low-rank local connectivity is a better inductive bias for highly aligned data while the input-dependent low rank local connectivity is better suited to less aligned datasets (Figure \ref{fig: fixed vs input dependent ranks}).

To further investigate this finding, we destroyed the alignment in CelebA by placing the $32\times32$ face images uniformly within a $48\times48$ image with random uniform noise, thus randomly translating the CelebA faces. Our results show that the LRLC accuracy on `Translated CelebA' dropped while input dependent LRLC accuracy remained largely invariant to the translation (Figure \ref{fig: translated celeba}). We further visualized the combining weights with models of rank 2 so that the results may be easily interpreted. Our results show that the combining weights for the LRLC layer uses one filter bank for central positions where translated faces overlap most, and the other for the periphery (Figure \ref{fig: visualization.} left). For the input dependent LRLC, the combining weights tracked the translated faces, which enables the layer to capture spatially varying information in less aligned datasets (Figure \ref{fig: visualization.} eight).

\subsection{Feasibility of application of low-rank local connectivity to large scale problems}
\label{sec:large scale}

\begin{table*}[t]
 \caption{\textbf{Application of LRLC layer to ImageNet} Top-1 accuracy of different Resnet-50 models (mean $\pm$ SE). The optimal rank for LRLC models is obtained by evaluating models on a separate validation subset.}
 \label{table:imagenet}
\vskip 0.1in
\begin{center}
\begin{small}
\begin{sc}
\begin{tabular}{lcccr}
\toprule
Layer & Insert Layer & Replace 3x3 convs \\
\midrule
CONVOLUTION & $77.22\pm0.03$ & $76.93\pm0.07$ \\
COORDCONV(LIU ET AL., 2018) & $77.23\pm0.03$ & $77.07\pm0.08$ \\
LRLC & $77.47\pm0.03$ & $77.08\pm0.02$ \\
LRLC (INPUT DEPENDENT WEIGHTS) & $77.45\pm0.03$ & $77.80\pm0.02$ \\
WIDE CONVOLUTION & \textbf{77.48$\pm$0.05} &  \textbf{78.54$\pm$0.04} \\
\bottomrule
\end{tabular}
\end{sc}
\end{small}
\end{center}
\vskip -0.1in
\end{table*}

In this section, we demonstrate the feasibility of using the low-rank locally connected layers in large scale problems. Locally connected layers are not suitable to large scale problems as the number of trainable parameters scale with spatial dimension, which can be prohibitively large in dataset with high-resolution images. For example, a locally connected layer applied to $224x \time 224$ images from ImageNet would require 50176 local filter banks in a locally connected layer. In contrast, the number of filter banks in the low-rank locally connected layers only scales with the rank parameter, which in practice is much smaller than the spatial dimensionality.

To demonstrate the feasibility of using the LRLC layers in practice, we conducted two experiments with ResNet-50 on ImageNet (see Appendix \ref{sec:imagenet training}
 for training details). In the first experiment, we inserted one additional LRLC layer after the first convolution layer. In the second experiment, we replaced all $3\times3$ convolutions in the network blocks with the LRLC layer. Note that these experiments would have been prohibitively expensive if we used a vanilla locally connected layer instead. We explored spatial ranks 1, 4 and 7 and picked the best rank using a holdout dataset split. Similar to the previous results in MNIST, CIFAR-10, and CelebA, the LRLC models outperformed convolution, which suggests that ImageNet also benefits from relaxing spatial invariance (Table \ref{table:imagenet}). However, on ImageNet a wider version of ResNet-50 which matches the number of parameters in LRLC either matches or outperforms LRLC (Table \ref{table:imagenet}). The feasibility of running these large scale experiments opens the door for the utilization of the LRLC layer in many computer vision problems.

%% file: conclusion.tex
\section{Conclusion}
\label{conclusion}

In this work, we tested whether spatial invariance, a fundamental property of convolutional layers, is an overly restrictive inductive bias.
To address this question, we designed a new locally connected layer (LRLC) where the degree of spatial invariance can be controlled by modifying a spatial rank parameter. This parameter determines the size of the basis set of local filter banks that the layer can use to form local filters at different locations of the input. The LRLC layer has a similar limitation to locally connected layers that it has more trainable parameters than convolution layers. However, the LRLC parameters' count scale only with the spatial rank, which is much smaller scaling compared to the scale by spatial dimensionality in locally connected layers.

Our results show that relaxing spatial invariance using our LRLC layer enhances the accuracy of models over standard convolutional networks, indicating that spatial invariance may be overly restrictive.
However, we also found that our proposed LRLC layer achieves higher accuracy than a vanilla locally connected layer, indicating that there are benefits to \textit{partial} spatial invariance.
We show that relaxing spatial invariance in later layers is better than relaxing spatial invariance in early layers.
Further, we find that the input dependent LRLC layer, which adapts local filters to each input, appears to perform better when data  are not well-aligned.

Locally connected layers have largely been ignored by the research community due to the perception that they perform poorly and the complexity of the number of their trainable parameter. However, our findings suggest that this pessimism should be reexamined, as locally connected layers with our low-rank parameterization achieve promising performance and solves the trainable parameters complexity problem. 
Further work is necessary to
capture the advantages of relaxing spatial invariance on other computer vision problems and datasets. One interesting direction to achieve this goal could be to utilize our LRLC formulation and explore using basis set with mixed filter sizes and dilation rates to construct a variety of layers that could suit datasets from different applications.

%% file: appendix.tex
\onecolumn
\appendix

\part*{Appendix}

\setcounter{figure}{0} \renewcommand{\thefigure}{Supp.\arabic{figure}}
\setcounter{table}{0} \renewcommand{\thetable}{Supp.\arabic{table}}
\newcommand{\underscore}{$\_$}

\section{Supplementary figures}
\label{sec: supp figures}

\begin{figure}[hbt!]
\begin{center}
\centerline{\includegraphics[width=\columnwidth]{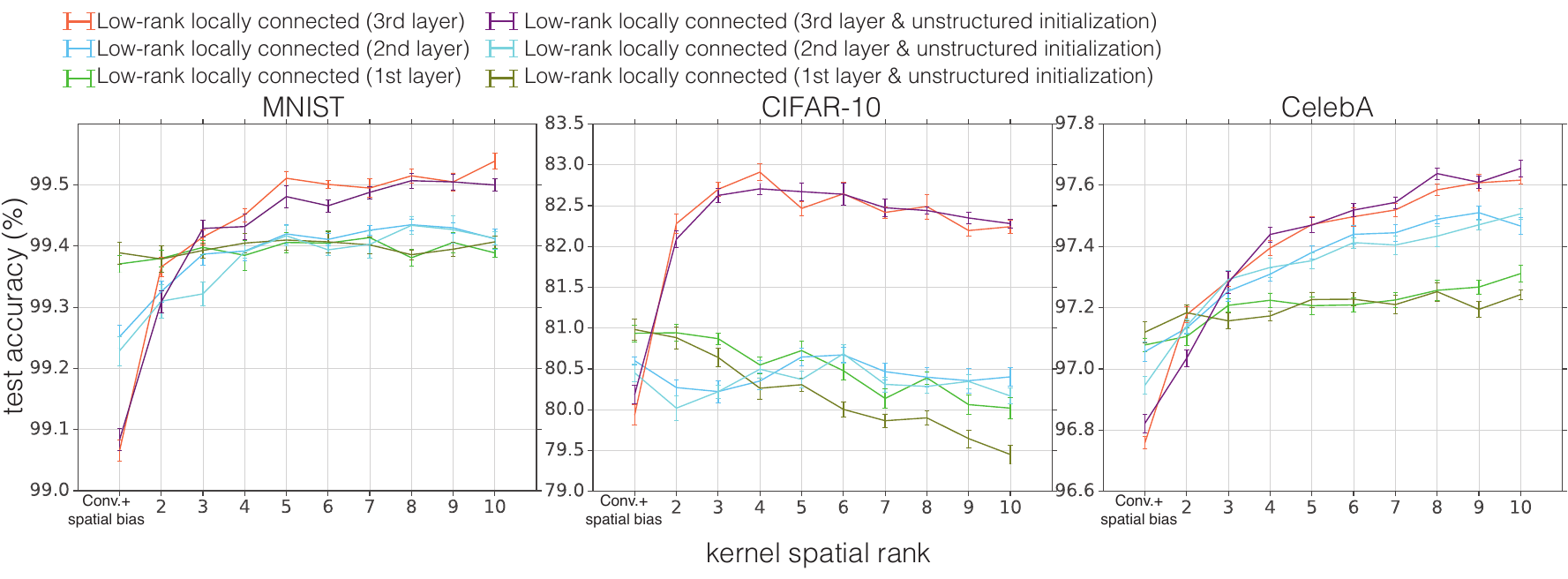}}
\caption{\textbf{Structured vs unstructured initialization.} Top 1 accuracy similar to Figure \ref{fig: lrlc}. We study the effect of the structured initialization  we used in our experiments for the LRLC layers (i.e., initialization to a convolution layer with a random kernel). In the structured initialization, we initialized the layer combining weights to constant equal to $1/\sqrt{\text{spatial rank}}$. We compared this initialization to a random initialization of the combining weights. Our results show that the structured initialization is generally quite similar to the unstructured initialization. Error bars indicate $\pm$ standard errors computed from training models from 10 different random initialization. 
}
\label{fig: structured initialization}
\end{center}
\vskip -0.2in
\end{figure}

\begin{figure}[hbt!]
\begin{center}
\centerline{\includegraphics[width=\columnwidth]{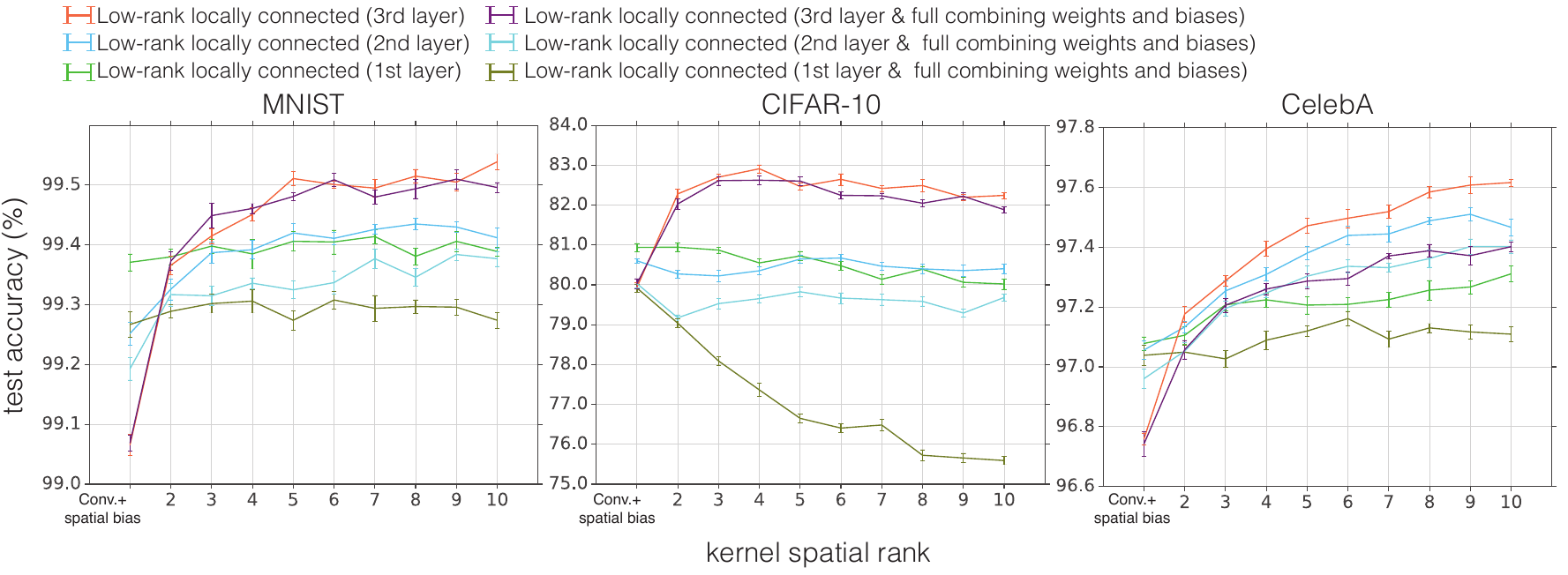}}
\caption{\textbf{Factorized vs full combining weights and biases.} Top 1 accuracy similar to Figure \ref{fig: lrlc}. We study the effect of factorizing the combining weights and biases in Equations \ref{eqn: combining weights} and \ref{eqn: biases}. We compare the performance of a LRLC layer with factorized weights and bias to a LRLC without this factorization. The layer with the factorization seems to perform better.}
\label{fig: factorized weights and biases}
\end{center}
\vskip -0.2in
\end{figure}

\begin{figure}[hbt!]
\begin{center}
\centerline{\includegraphics[width=\columnwidth]{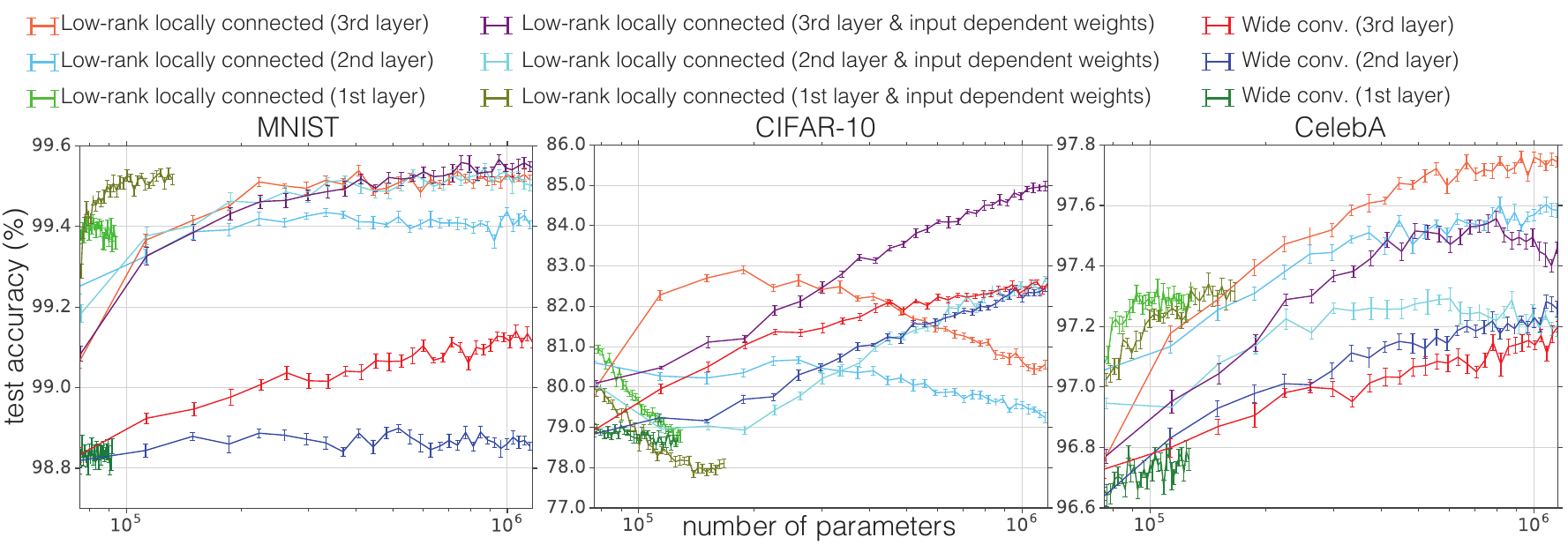}}
\caption{\textbf{Accuracy as a function of model parameters.} Classification accuracy as a function of network parameters. Error bars indicate $\pm$ standard errors computed from training models from 10 different random initialization.}
\label{fig: fixed vs input dependent params}
\end{center}
\vskip -0.2in
\end{figure}

\begin{figure}[hbt!]
\begin{center}
\centerline{\includegraphics[width=0.5\columnwidth]{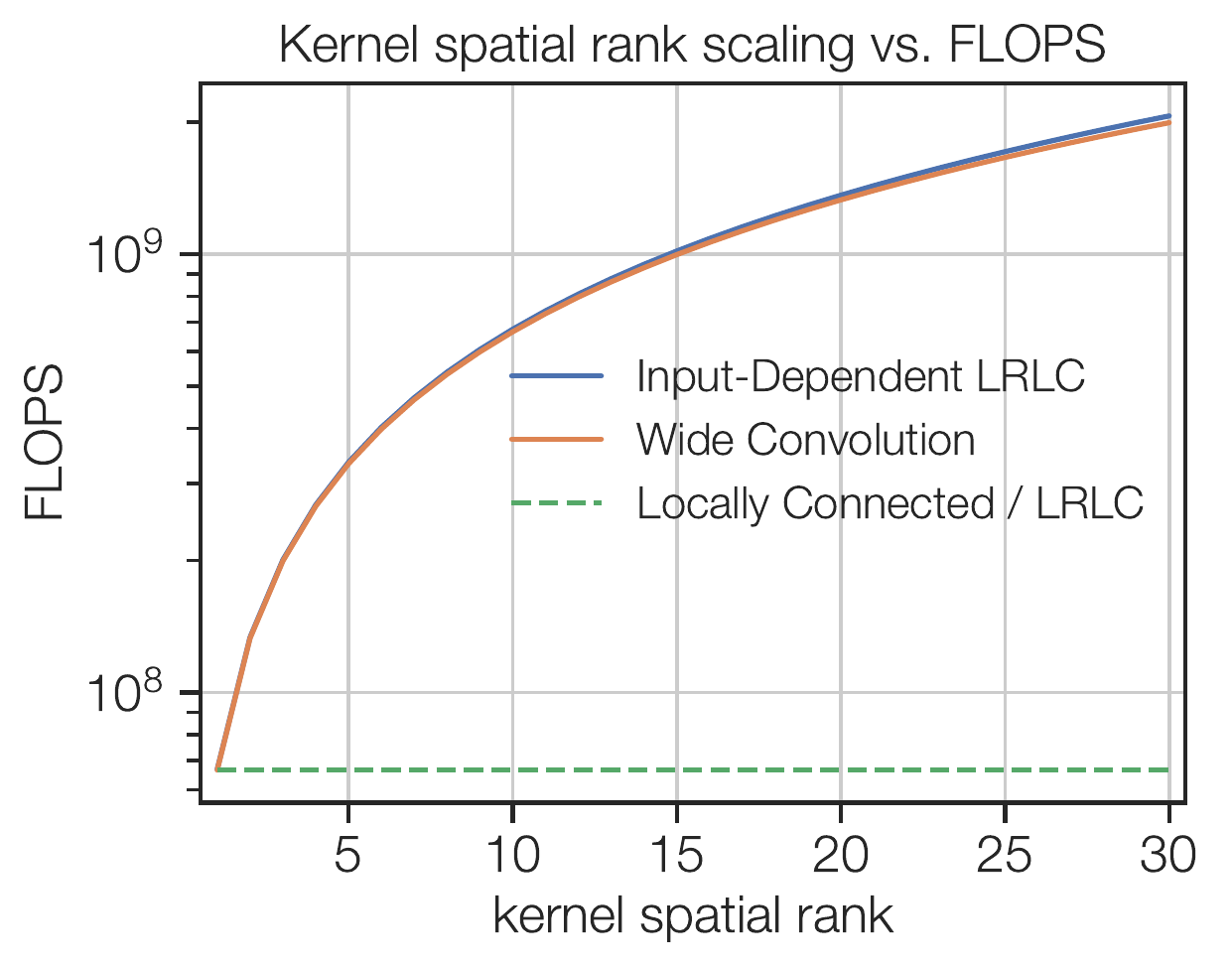}}
\caption{\textbf{Computational cost as function of the spatial rank of the locally connected kernel.} As the spatial rank of the locally connected kernel increases, the computational cost, as measured by the number of floating point operations (FLOPS), of the input-dependent LRLC layer and the convolution layer with similar trainable parameter (wide convolution) grows at a similar rate, while the computational cost of the LRLC layer stays constant because it can be converted into a locally connected layer at inference time.
}
\label{fig: lrlc flops}
\end{center}
\vskip -0.2in
\end{figure}

\begin{figure}[hbt!]
\begin{center}
\centerline{\includegraphics[width=\columnwidth]{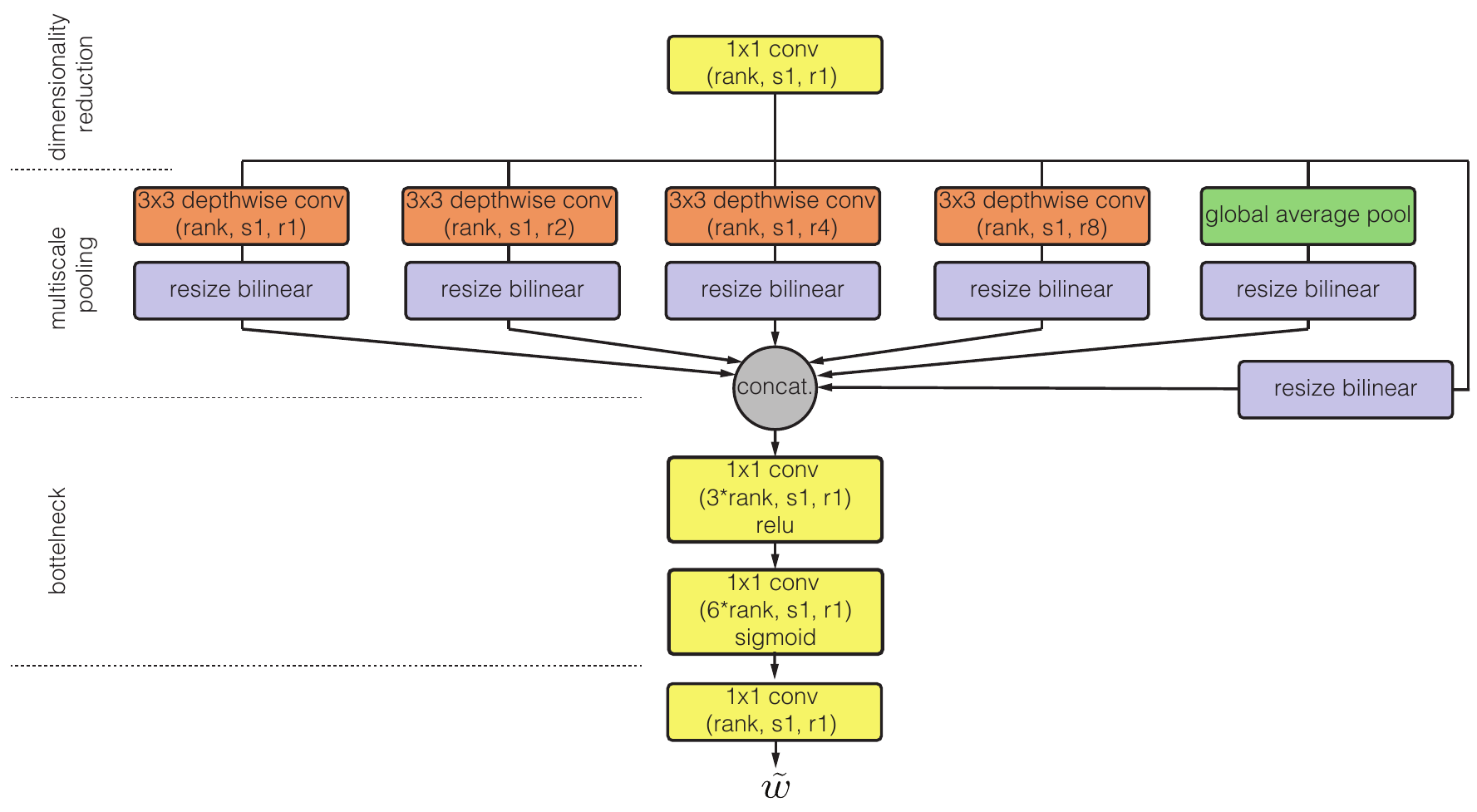}}
\caption{\textbf{Input-dependent combining weights network architecture.}}
\label{fig: input-dependent network}
\end{center}
\vskip -0.2in
\end{figure}

\begin{figure}[hbt!]
\begin{center}
\centerline{\includegraphics[width=\columnwidth]{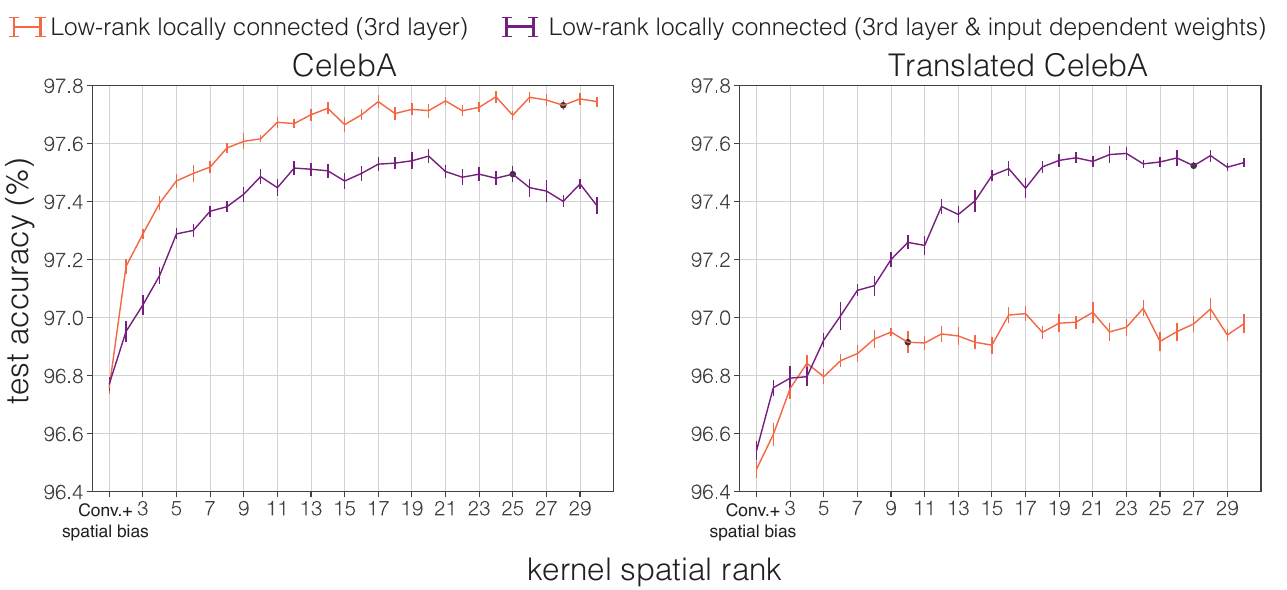}}
\caption{\textbf{Input-dependent LRLC is invariant to translation.} Comparing the performance of LRLC and input dependent LRLC networks in CelebA and Translated CelebA datasets.}
\label{fig: translated celeba}
\end{center}
\vskip -0.2in
\end{figure}

\begin{figure}[hbt!]
\begin{center}
\centerline{\includegraphics[width=\columnwidth]{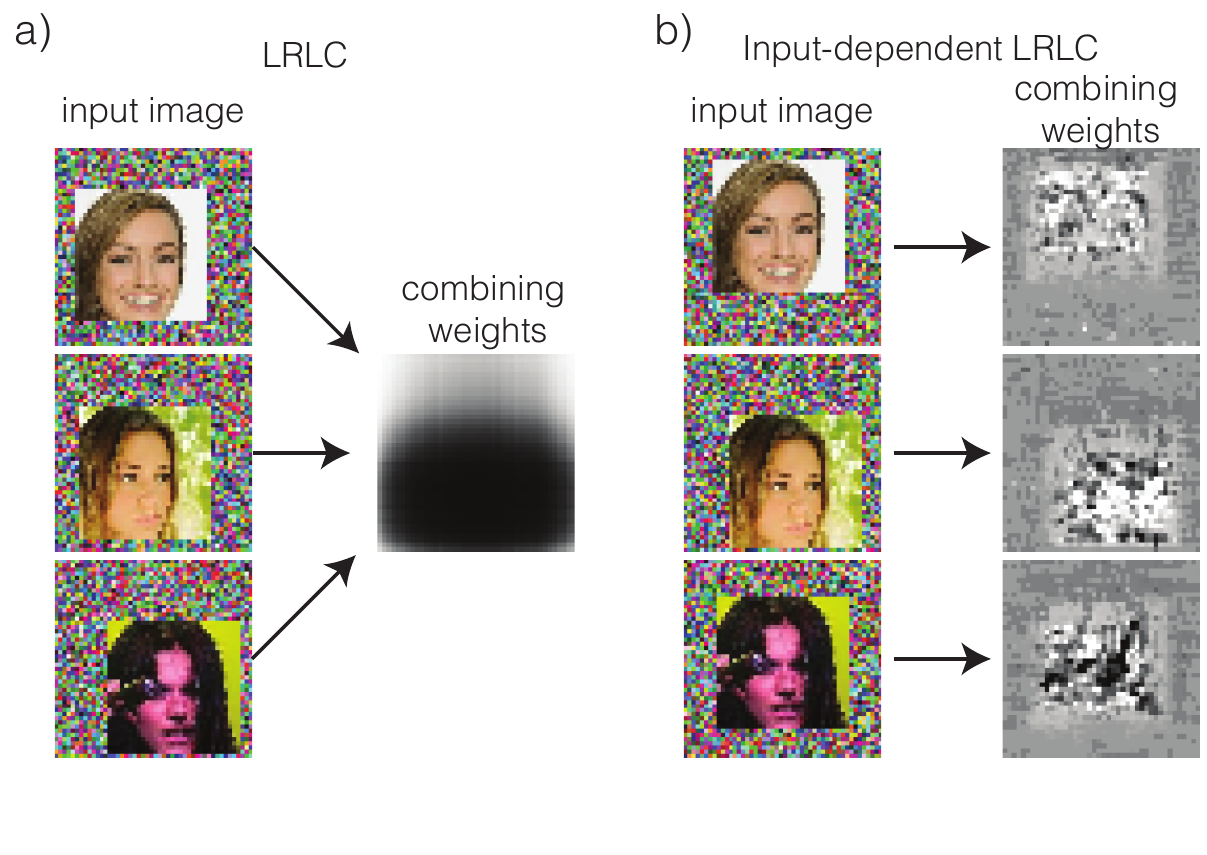}}
\caption{\textbf{Visualization of combining weights.} Combining weights for an LRLC network a) and input-dependent LRLC network in b) with rank 2 trained on Translated CelebA dataset.}
\label{fig: visualization.}
\end{center}
\vskip -0.2in
\end{figure}

\clearpage
\section{Input-dependent combining weights network}
\label{sec:input-dependent-kernel-network}
The architecture of the input-dependent combining weights network ($g$) is illustrated in Figure \ref{fig: input-dependent network}. The initial operation of $g$ is to project the input channels to a low-dimensional space using a $1 \times 1$ convolution. This projection is used to allow $g$ to have small number of parameters, and also because selection of filter banks in the basis set is potentially a simpler task than the classification task the network is performing. Motivated by work on segmentation  \cite{chen2017deeplab, yu2015multi,chen2017rethinking}, the second operation collects statistics across different scales of the input using parallel pooling and dilated depth-wise $3 \times 3$ convolution layers followed by bilinear resizing. Note the increase in parameters here is small due to the initial projection step and the use of depth-wise convolution. The next stage is a nonlinear low-dimensional bottleneck followed by nonlinear dimensionality expansion with $1 \times 1$ convolution. This operation has similar flavor to the Squeeze-and-Excitation operation \cite{Hu_2018_CVPR}, and is included to give $g$ the power to learn useful embedding of the input. The last layer is a linear $1 \times 1$ convolution that reduce the channels size to the spatial rank.

\section{ImageNet training}
\label{sec:imagenet training}
We divide the standard ImageNet ILSVRC 2012 training set into training and development subsets. We trained our models on the training subset and chose best rank based on the development subset. We follow common practice and report results on the separate ILSVRC 2012 validation set, which we do not use for training or hyperparameter selection. 

We trained the network by optimizing the cross entropy loss plus $\ell^2$-regularization on the model weights. We optimized all models using Stochastic Gradient Descent with Nesterov momentum of 0.9. We preprocessed images by subtracting the mean and dividing by the standard deviation of training examples. During optimization, we augmented the training data by taking a random crop within the image and then performing bicubic resizing to model’s resolution. We used a batch size of 2048 and $\ell^2$-regularization scale of $8e-5$. We trained our models for 150 epochs starting with a linear warmup period of 10 epochs and used a cosine decay schedule afterwards. We used Tensor Processing Unit (TPU) accelerators in all our training. We computed our results by computing the top-1 accuracy $\pm$ standatd error based on models trained from 3 different random initializations.

\clearpage
\section{Supplementary tables}
\label{sec: supp tables}

\begin{table*}[!h]
\caption{\textbf{Number of examples in each dataset.}}
\label{table: data subsets}
\vskip 0.1in
\begin{center}
\begin{small}
\begin{sc}
\begin{tabular}{lcccr}
\toprule
subset & MNIST & CIFAR-10 & CELEBA \\
\midrule
Train    & 55000 &45000& 162770  \\
Validation &5000 & 5000& 19867 \\
Test    &10000  &10000  & 19962 \\
\bottomrule
\end{tabular}
\end{sc}
\end{small}
\end{center}
\vskip -0.1in
\end{table*}

\begin{table*}[!h]
 \caption{\textbf{Summary of results (train subset).} Top-1 train accuracy of different models (mean $\pm$ SE).}
 \label{table: train accuracy}
\vskip 0.1in
\begin{center}
\begin{small}
\begin{sc}
\begin{tabular}{lcccr}
\begin{filecontents*}{figures/results_table_top1_train.csv}
Layer,MNIST,CIFAR-10,CelebA
Convolution,$100.00\pm0.00$,$97.47\pm0.07$,$100.00\pm0.00$
Convolution (fc),$100.00\pm0.00$,$100.00\pm0.00$,$100.00\pm0.00$
Convolution + spatial bias (1st layer),$100.00\pm0.00$,$97.99\pm0.04$,$100.00\pm0.00$
Convolution + spatial bias (2nd layer),$100.00\pm0.00$,$97.99\pm0.06$,$100.00\pm0.00$
Convolution + spatial bias (3rd layer),$100.00\pm0.00$,$97.60\pm0.05$,$100.00\pm0.00$
Convolution + spatial bias (all layers),$100.00\pm0.00$,$98.36\pm0.06$,$100.00\pm0.00$
CoordConv,$100.00\pm0.00$,$97.30\pm0.08$,$100.00\pm0.00$
LRLC (1st layer \& input dependent weights),$100.00\pm0.00$,$99.04\pm0.05$,$100.00\pm0.00$
LRLC (1st layer),$100.00\pm0.00$,$98.53\pm0.04$,$100.00\pm0.00$
LRLC (2nd layer \& input dependent weights),$100.00\pm0.00$,$100.00\pm0.00$,$100.00\pm0.00$
LRLC (2nd layer),$100.00\pm0.00$,$100.00\pm0.00$,$100.00\pm0.00$
LRLC (3rd layer \& input dependent weights),$100.00\pm0.00$,$100.00\pm0.00$,$100.00\pm0.00$
LRLC (3rd layer),$100.00\pm0.00$,$100.00\pm0.00$,$100.00\pm0.00$
LRLC (all layers),$100.00\pm0.00$,$100.00\pm0.00$,$100.00\pm0.00$
Locally connected (1st layer),$100.00\pm0.00$,$100.00\pm0.00$,$100.00\pm0.00$
Locally connected (2nd layer),$100.00\pm0.00$,$100.00\pm0.00$,$100.00\pm0.00$
Locally connected (3rd layer),$100.00\pm0.00$,$100.00\pm0.00$,$100.00\pm0.00$
Wide convolution (1st layer),$100.00\pm0.00$,$98.25\pm0.05$,$100.00\pm0.00$
Wide convolution (2nd layer),$100.00\pm0.00$,$100.00\pm0.00$,$100.00\pm0.00$
Wide convolution (3rd layer),$100.00\pm0.00$,$100.00\pm0.00$,$100.00\pm0.00$
\end{filecontents*}
\csvautobooktabular{figures/results_table_top1_train.csv}
\end{tabular}
\end{sc}
\end{small}
{\small \\CONVOLUTION (FC) is a convolution network with a fully connected last layer and without global average pooling.}
\end{center}
\vskip -0.1in
\end{table*}

\begin{table*}[!h]
 \caption{\textbf{Summary of results (validation subset).} Top-1 test accuracy of different models (mean $\pm$ SE).}
 \label{table: test accuracy}
\vskip 0.1in
\begin{center}
\begin{small}
\begin{sc}
\begin{tabular}{lcccr}
\begin{filecontents*}{figures/results_table_top1_valid.csv}
LAYER,MNIST,CIFAR-10,CelebA
Convolution,$98.89\pm0.03$,$79.68\pm0.13$,$97.27\pm0.02$
Convolution (fc),$98.79\pm0.03$,$68.36\pm0.17$,$97.48\pm0.02$
Convolution + spatial bias (1st layer),$99.28\pm0.02$,$81.61\pm0.14$,$97.61\pm0.02$
Convolution + spatial bias (2nd layer),$99.16\pm0.02$,$81.08\pm0.09$,$97.56\pm0.02$
Convolution + spatial bias (3rd layer),$98.99\pm0.02$,$80.83\pm0.11$,$97.28\pm0.05$
Convolution + spatial bias (all layers),$99.25\pm0.02$,$81.29\pm0.12$,$97.59\pm0.02$
CoordConv,$99.33\pm0.02$,$81.97\pm0.15$,$97.83\pm0.02$
LRLC (1st layer \& input dependent weights),$99.36\pm0.02$,$80.50\pm0.11$,$97.82\pm0.03$
LRLC (1st layer),$99.30\pm0.02$,$81.76\pm0.11$,$97.79\pm0.02$
LRLC (2nd layer \& input dependent weights),$99.35\pm0.03$,$83.55\pm0.14$,$97.82\pm0.02$
LRLC (2nd layer),$99.31\pm0.02$,$81.18\pm0.10$,$98.03\pm0.03$
LRLC (3rd layer \& input dependent weights),$99.39\pm0.02$,$85.61\pm0.06$,$98.00\pm0.03$
LRLC (3rd layer),$99.41\pm0.01$,$82.87\pm0.13$,$98.16\pm0.02$
LRLC (all layers),$99.31\pm0.02$,$81.75\pm0.10$,$97.98\pm0.02$
Locally connected (1st layer),$98.77\pm0.02$,$63.72\pm0.20$,$96.95\pm0.01$
Locally connected (2nd layer),$98.69\pm0.02$,$70.24\pm0.15$,$97.63\pm0.02$
Locally connected (3rd layer),$99.00\pm0.01$,$72.49\pm0.14$,$97.79\pm0.02$
Wide convolution (1st layer),$98.91\pm0.02$,$79.94\pm0.14$,$97.37\pm0.02$
Wide convolution (2nd layer),$98.91\pm0.02$,$82.98\pm0.06$,$97.90\pm0.02$
Wide convolution (3rd layer),$99.13\pm0.02$,$83.11\pm0.07$,$97.72\pm0.02$
\end{filecontents*}
\csvautobooktabular{figures/results_table_top1_valid.csv}
\end{tabular}
\end{sc}
\end{small}
{\small \\CONVOLUTION (FC) is a convolution network with a fully connected last layer and without global average pooling.}
\end{center}
\vskip -0.1in
\end{table*}

\begin{table*}[!h]
 \caption{\textbf{Summary of results (test subset).} Top-1 test accuracy of different models (mean $\pm$ SE). The optimal rank in LRLC and the optimal width in wide convolution models are obtained by evaluating models on a separate validation subset.}
 \label{table: test accuracy}
\vskip 0.1in
\begin{center}
\begin{small}
\begin{sc}
\begin{tabular}{lcccr}
\begin{filecontents*}{figures/results_table_top1_test.csv}
Layer,MNIST,CIFAR-10,CelebA
Convolution,$98.84\pm0.01$,$78.80\pm0.12$,$96.66\pm0.04$
Convolution (fc),$99.05\pm0.02$,$67.59\pm0.19$,$96.98\pm0.02$
Convolution + spatial bias (1st layer),$99.37\pm0.01$,$80.94\pm0.10$,$97.08\pm0.02$
Convolution + spatial bias (2nd layer),$99.25\pm0.02$,$80.60\pm0.05$,$97.06\pm0.03$
Convolution + spatial bias (3rd layer),$99.07\pm0.02$,$79.94\pm0.12$,$96.76\pm0.02$
Convolution + spatial bias (all layers),$99.36\pm0.01$,$80.89\pm0.08$,$97.12\pm0.03$
CoordConv,$99.46\pm0.01$,$81.29\pm0.13$,$97.40\pm0.02$
LRLC (1st layer \& input dependent weights),$99.52\pm0.02$,$79.99\pm0.07$,$97.32\pm0.03$
LRLC (1st layer),$99.39\pm0.02$,$80.94\pm0.10$,$97.25\pm0.02$
LRLC (2nd layer \& input dependent weights),$99.53\pm0.01$,$82.62\pm0.07$,$97.25\pm0.02$
LRLC (2nd layer),$99.42\pm0.02$,$80.67\pm0.09$,$97.53\pm0.01$
LRLC (3rd layer \& input dependent weights),$\boldsymbol{99.57\pm0.01}$,$\boldsymbol{84.91\pm0.07}$,$97.49\pm0.03$
LRLC (3rd layer),$99.51\pm0.01$,$82.70\pm0.08$,$\boldsymbol{97.73\pm0.02}$
LRLC (all layers),$99.45\pm0.01$,$81.03\pm0.11$,$97.51\pm0.03$
Locally connected (1st layer),$98.74\pm0.01$,$62.54\pm0.14$,$96.32\pm0.02$
Locally connected (2nd layer),$98.72\pm0.02$,$69.29\pm0.09$,$97.19\pm0.02$
Locally connected (3rd layer),$99.10\pm0.01$,$71.86\pm0.10$,$97.31\pm0.01$
Wide convolution (1st layer),$98.84\pm0.02$,$78.96\pm0.11$,$96.77\pm0.04$
Wide convolution (2nd layer),$98.89\pm0.02$,$82.36\pm0.09$,$97.28\pm0.02$
Wide convolution (3rd layer),$99.10\pm0.02$,$82.40\pm0.10$,$97.20\pm0.02$
\end{filecontents*}
\csvautobooktabular{figures/results_table_top1_test.csv}
\end{tabular}
\end{sc}
\end{small}
{\small \\CONVOLUTION (FC) is a convolution network with a fully connected last layer and without global average pooling.}
\end{center}
\vskip -0.1in
\end{table*}